\documentclass{article}
\usepackage{fullpage}
\usepackage{natbib}
\usepackage{graphicx}

\title{\bf Velocity variations at Columbia Glacier captured by particle filtering of oblique time-lapse images}
\author{{\bf Douglas Brinkerhoff}\thanks{Correspondence to: douglas1.brinkerhoff@umontana.edu}\\
        University of Montana \and 
        \bf{Shad O'Neel}\\
        United States Geological Survey}
\date{}

\begin{document}
\maketitle

\begin{abstract}
We develop a probabilistic method for tracking glacier surface motion based on time-lapse imagery, which works by sequentially resampling a stochastic state-space model according to a likelihood determined through correlation between reference and test images.  The method is robust due to its natural handling of periodic occlusion and its capacity to follow multiple hypothesis displacements between images, and can improve estimates of velocity magnitude and direction through the inclusion of observations from an arbitrary number of cameras.  We apply the method to an annual record of images from two cameras near the terminus of Columbia Glacier.  While the method produces velocities at daily resolution, we verify our results by comparing eleven-day means to TerraSar-X.  We find that Columbia Glacier transitions between a winter state characterized by moderate velocities and little temporal variability, to an early summer speed-up in which velocities are sensitive to increases in melt- and rainwater, to a fall slowdown, where velocities drop to below their winter mean and become insensitive to external forcing, a pattern consistent with the development and collapse of efficient and inefficient subglacial hydrologic networks throughout the year.  
\end{abstract}

\section{Introduction}
Motion is what defines a glacier, and measuring this motion is a principal concern for understanding changing ice dynamics.  Observed over multiple years, ice velocity and acceleration inform the dynamic component of glacial contributions to sea level rise.  \citep[e.g.][]{Burgess:2013,Joughin:2010}.  In the very short term (e.g. hourly measurement intervals), velocity measurements can elucidate the physics of glacial response to diurnal or tidal forcing \citep{Dietrich:2007,Meier:1993}.  At intermediate scales, velocity changes provide information on the sensitivity of glacier flow to changes in liquid input (i.e. from storms or hot days) and on the configuration of the subglacial drainage system that determines this sensitivity \citep{Harper:2007}.  Analysis of image sequences captured by ground-based cameras provides a compelling mechanism for evaluating glacier velocity variations at short and intermediate time scales because of the low cost and ubiquity of the necessary equipment.  While many photogrammetric techniques are applicable to both orthogonal and oblique imagery, we confine our review to the latter.  At the most basic level, methods of capturing the velocity consist of two elements: first, a means of tracking persistent features in image coordinates.  Second, a relation between image and spatial coordinates.  

The first quantitative application of time-lapse imagery to glaciers in was by \citet{Flotron:1973}, who resolved both the seasonal signal of horizontal velocity, but also a small signal of vertical motion.  In Alaska, \citet{Harrison:1986} used automatic 35mm cameras to document daily speedups on Variegated Glacier during the three summers prior to its surge in 1982.  \citet{Krimmel:1986} used a similar system at the terminus to Columbia Glacier during the early stages of its retreat, manually tracking 30 points through a sequence of 3 frames per day of the summer of 1983.  \citet{Harrison:1992} describe an approach to tackling one of the more significant difficulties associated with tracking slow moving fluid: the camera also has a tendency to move and that control points are not always available.  Despite the lack of surveyed ground control points or a stable camera position, they were able to document the surge of West Fork Susitna Glacier.

Since the advent of digital cameras, a handful of methods have emerged by which to perform automatic measurements with oblique cameras, in contrast to the manual methods used previously.  \citet{Evans:2000} used a probabilistic metric to track multiple potential flow pathways.  However, he stopped short of converting solutions from image coordinates to spatial coordinates, and so the utility of this tool for practical glaciology is limited.  \citet{Dietrich:2007} used automated tracking of features on the surface of Jakobshavn Isbr\ae, in conjunction with a photogrammetrically-derived DEM, resolving both horizontal and vertical motion of the Jakobshavn terminus under an \emph{a priori} assumption of flow direction from a remotely-sensed velocity field.  Cameras were calibrated and corrected with surveyed fiduciary points, and were oriented such that they were orthogonal to the primary flow direction.  Their work represents best practices in camera installation, but these were usually not considered for cameras where feature tracking was not the initial intent.  \citet{Rosenau:2013} expanded this experiment over several more years, localizing the grounding line position at Jakobshavn based on presence or absence of tidal motion.  At Helheim Glacier, feature tracking was similarly used to quantify tidal flexure near the glacier terminus \citep{Murray:2015,James:2014}.  \citet{Ahn:2010} demonstrated a robust method for tracking flow fields utilizing multiple image pre-processing techniques simultaneously to reduce the incidence of incorrect matches between image pairs from several cameras in Greenland.  The resulting velocities agree well with those derived from remote sensing.  Finally, \citet{Messerli:2014} and \citet{James:2016} provided software libraries that distill many of the essential methods required for oblique time-lapse velocity measurements into accessible and open-source packages.

In this paper, we address the problem of determining glacier velocities from an unstable camera under changing lighting conditions without precise ground control.  In contrast to previous work, we apply a probabilistic approach that allows us to address some of the key difficulties of other methods while producing robust estimates of uncertainty under unfavorable conditions that are still ubiquitous 25 years after \citet{Harrison:1992}.  We apply a method called particle filtering, which sequentially updates the probability distribution of the state of a moving glacier surface, namely its position and velocity by considering a likelihood derived from matching characteristic features between images.  Besides robust error accounting, this method immediately generalizes to multiple cameras, is robust to partial occlusion due to the use of predictions by an underlying physical model, and does not require the ill-posed step of projecting values in image coordinates onto the landscape.  We apply the method to images collected by a pair of (non-stereo) cameras at Columbia Glacier, AK, that are subject to all of the potential uncertainty sources typical of time-lapse image analysis.  We process images over nearly a full year, for the first time providing speeds over all seasons at daily temporal resolution, providing new insight into the evolution of basal processes near the terminus of a large tidewater glacier.

\section{Columbia Glacier}
\begin{figure*}
\includegraphics[width=\linewidth]{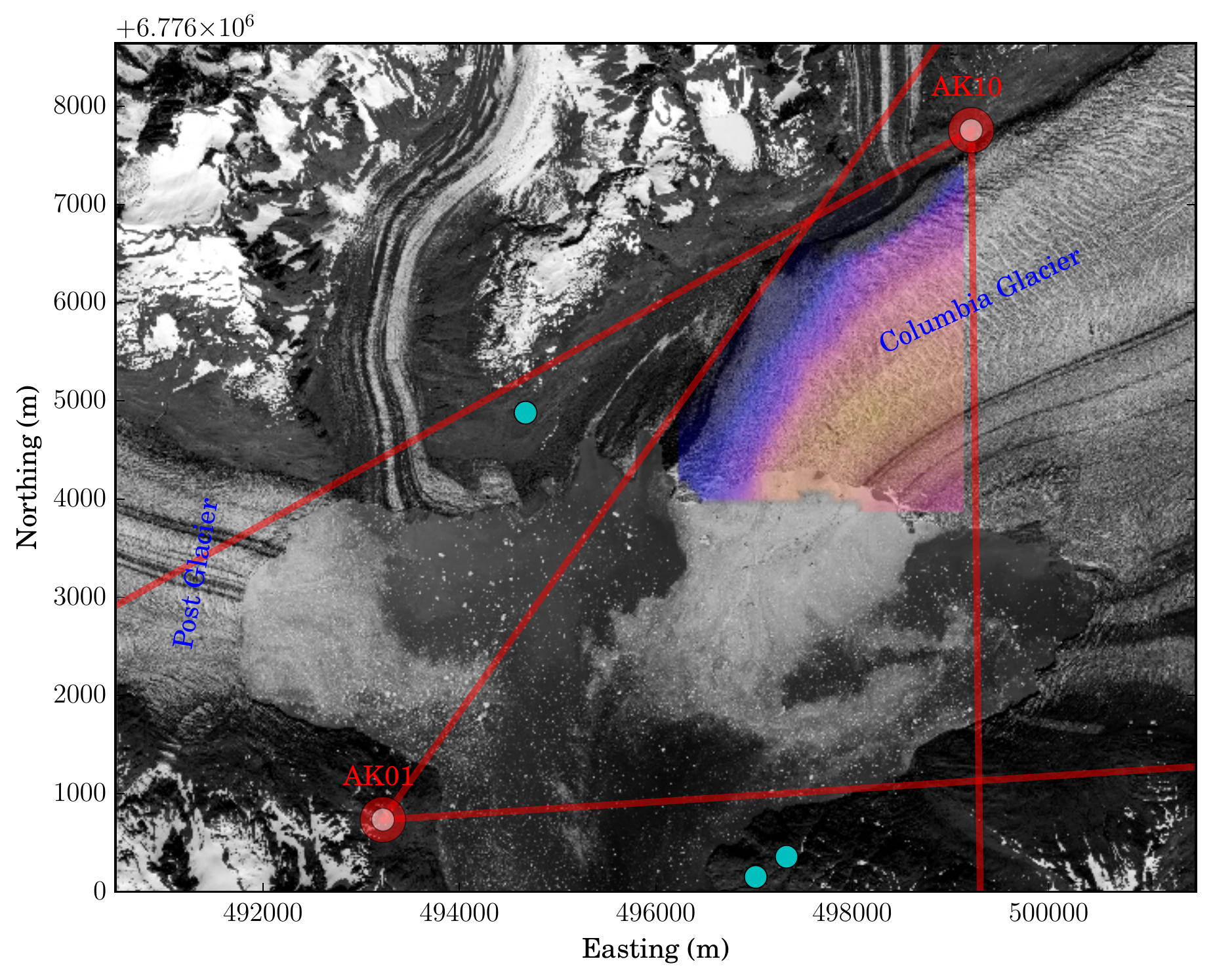}
\caption[Columbia Glacier camera configuration and map.]{Columbia Bay and the Columbia Glacier terminus c. July 2013.  Cameras used in this study are denoted by red bullseyes, and their field of view by red lines.  Additional time-lapse cameras that were not included are given by cyan circles.  We compute velocity fields in the glacierized region that falls within the field of view of both cameras.  The colored surface is the glacier speed on 2013.07.15 (see Fig.~\ref{fig:images} for colorbar).  The temperature data shown in Fig.~\ref{fig:time_series} is recorded at a station $\sim$2 km out of frame to the northeast.  Base image courtesy Polar Geospatial Center.}
\label{fig:columbia}
\end{figure*}
\begin{figure*}
\includegraphics[width=\linewidth]{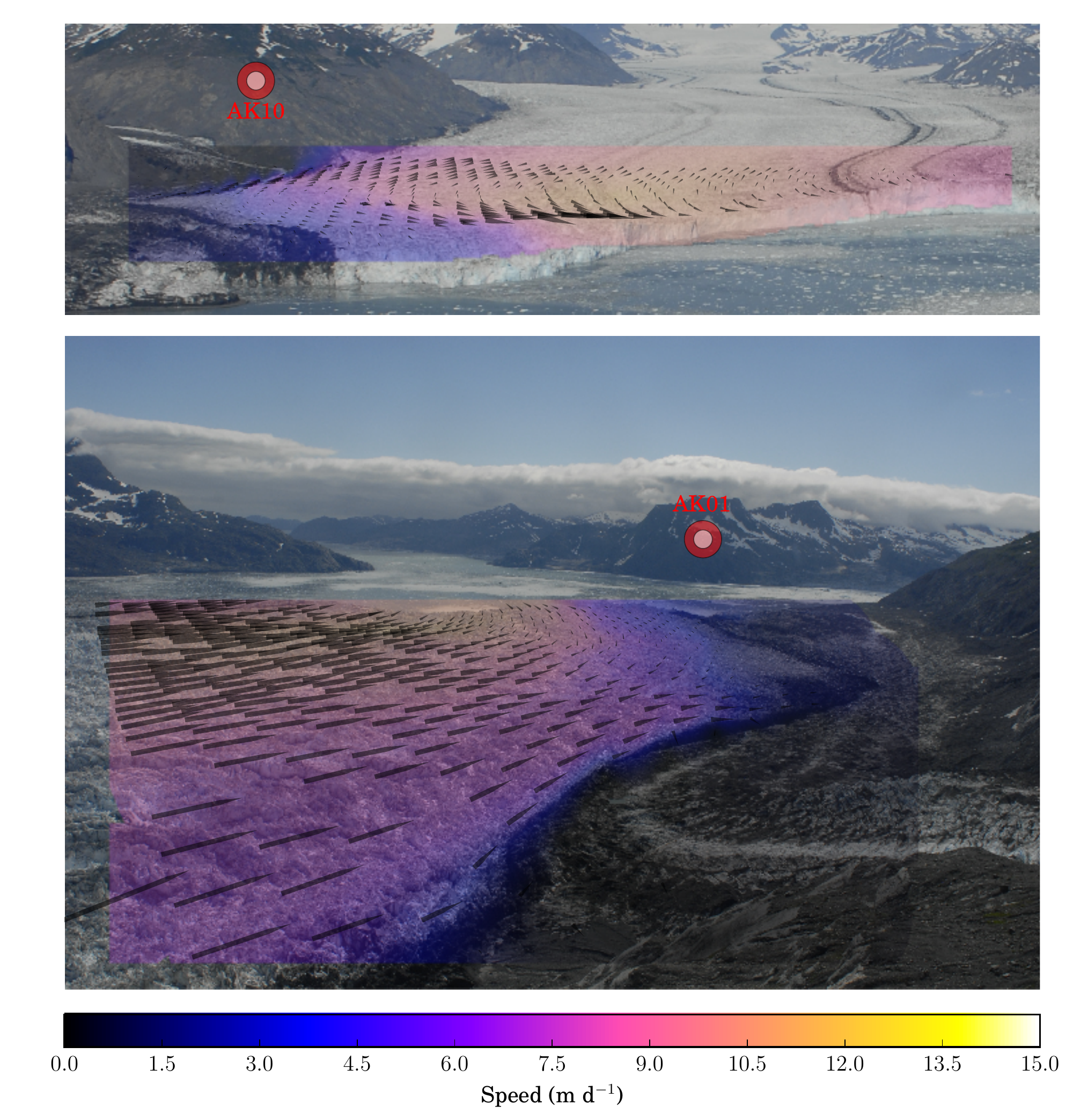}
\caption[Representative images for AK01 and AK10.]{Representative images for AK01 (top) and AK10 (bottom), the two cameras used in this study.  The location of the other camera is denoted by a bullseye.  The colors represent glacier speed on 2013.07.15, with vectors indicating flow speed and direction in image coordinates.  Note the presence of a strong shear margin which is well captured by the algorithm, as well as the very low image aspect ratios.}
\label{fig:images}
\end{figure*}
Columbia Glacier (Fig.~\ref{fig:columbia}a) is a 52 km long tidewater glacier that drains the high central Chugach Mountains in southcentral Alaska.  Its climate is strongly maritime with annual precipitation rates of near 3m at the terminus.  The terminus itself is characterized by high speeds and vigorous calving \citep{Meier:1987}.  Winter temperatures are moderate, while early summers see strong melt.  large rainstorms are common in the late summer and fall \citep{Bieniek:2012}.  

Columbia Glacier is currently in the retreat phase of the tidewater glacier cycle \citep{Meier:1987}, and has retreated more than 25 km from its maximum in 1980.  The specific mechanism for initiating the retreat is debatable \citep{Sikonia:1980,Carlson:2017}, as the glacier geometry at Columbia Glacier's maximum length was highly unstable due to the significant overdeepening in what is now Columbia's fjord.  In any case, the current retreat is driven by glacial dynamics and bed topography, with climate assuming an ancillary role \citep{Pfeffer:2007}.    

The velocity of Columbia Glacier, particularly near the terminus, has varied greatly since the beginning of the retreat and these variations have been documented over a variety of temporal and spatial scales since the 1980s.  \citet{Vaughn:1985} used an automated laser rangefinder and a small set of reflectors installed in the ice near the terminus to measure pointwise velocities at 15 minute intervals.  Despite the excellent temporal resolution, this method was only applied to a handful of points, and the record only lasts for approximately 30 days during the summers of 1984-1986.  \citet{Meier:1993} performed similar measurements in 1987.  \citet{Krimmel:1986} performed one of the first examples of using oblique time-lapse photography to measure glacier velocities at daily resolution over both winter and summer.  The camera, situated over 5 km from the study site was able to produce daily offsets with a nominal precision of 1m d$^{-1}$.     

These temporally detailed results are supplemented with distributed velocity fields determined from repeat georectified aerial photographs with intervals of around two months \citep{Krimmel:2001}, in which characteristic surface features such as crevasses were manually identified in image pairs.  Since that time, major improvements in the availability of satellite imagery have allowed the automated generation of velocity fields with full coverage and enhanced temporal resolution.  \citet{Fahnestock:2016}, used optical imagery from Landsat8 to produce velocity fields for each 15-day offset image pair in the satellite's brief record, excluding periods when the landscape was obscured by clouds.  Circumventing the occlusion issues of optical imagery, \citet{Burgess:2013} used synthetic aperture radar (SAR) to generate velocity fields at Columbia Glacier (and elsewhere) with a temporal resolution of 46 days at irregular intervals between 2007 and 2009.  \citet{Joughin:2010} and \citet{Vijay:2017} independently used SAR observations with a higher return frequency to produce 11-day average velocities between 2010 and 2016, with approximately monthly frequency.  As a side note, \citet{Vijay:2017} produced commensurate digital elevation models that coincide with their velocity fields which are publicly available.

Taken in aggregate, a consistent story about the spatial and temporal variations in Columbia's flow has emerged, superimposed on tidewater retreat.  At seasonal time scales, Columbia has a velocity minimum in October or November, before a slow ramping up to more consistent values through mid-winter into early spring.  In late spring, the glacier rapidly accelerates, reaching a velocity maximum in May or June, before a decline back to the minimum state in the fall, a pattern that has persisted from when it was first enumerated \citep{Meier:1987} through present \citep{Vijay:2017}, albeit with a drift towards respective maxima and minima occurring later in the season.  The magnitude of these variations have changed throughout the course of the retreat.  In the early 1980s, maximum velocities were less than 10m d$^{-1}$, increased to 30 m d$^{-1}$ during the periods of most vigorous retreat in the mid 1990s to 2000s \citep{Oneel:2005}, and have been on the order of 15 m d$^{-1}$ since 2010 \citep{Vijay:2017}.  

Since ice geometry changes relatively slowly, and these velocities are an order of magnitude higher than could be explained by deformation, seasonal evolution of subglacial pressure is commonly thought to drive these variations due to the ability of water pressure to partially offset the normal stress exerted by ice on the bed and thus reducing friction.  Pressure variations are induced by changes in the availability of surface water (either melt or rainfall) or changes in the character of the subglacial drainage system \citep[e.g.][]{Iken:1997,Werder:2013}.  Based on water pressure observations from terrestrial glaciers, it is hypothesized that during winter the lack of surface inputs lead to a weakly-connected drainage system with moderate pressure that produce moderate velocities \citep{Iken:1997,Truffer:2006}.  At the onset of the melt season in spring, the additional water overloads this system, leading to high pressures and velocities, but also causing an efficient drainage network to form.  In the fall, as water input decreases, water pressure drops below the annual average before the drainage system can once again equilibrate to winter conditions.  

The multiple field campaigns of the 1980s reported velocity fluctuations superimposed on these longer term signals during summer.  Signals of acceleration were observed at both diurnal and tidal periods, presumably associated with short term changes in water pressure.  Additionally, stochastic events such as foehn winds or large rainstorms were also observed to cause complex increases in velocity, often (but not always) followed by a decrease to below pre-event speeds.  Interestingly, contemporaneous measurements of pressure and speed did not reveal a clear relationship between the two \citep{Kamb:1993}.  However, these velocity changes were well (but not perfectly) correlated with changes in water storage as inferred from proxies for influx and outflux \citep{Fahnestock:1991}.  The observed mismatch in the direct pressure signal was ascribed to heterogeneity in the local subglacial drainage system relative to the area-averaged value relevant to glacier dynamics and aliased by total water storage.

Since the observations described above, few direct observations of Columbia Glacier's velocity have been collected.  While remote sensing is extremely useful for describing synoptic features, it lacks the temporal resolution necessary for assessing glaciological response to short term changes in geometry (i.e. calving) or hydrology.  Given the remarkable changes that have occurred over the last 30 years, it is not clear whether Columbia Glacier's dynamics resemble those observed c. 1987.  Fortunately, Columbia Glacier has been the focus of an intense campaign of oblique time-lapse photography since 2007 as part of an Extreme Ice Survey and USGS monitoring program.  Since 2007, cameras at seven sites have been active, generally taking images of the glacier terminus at sub-hourly intervals (Fig.~\ref{fig:columbia}b).  Here we apply the method described below to these images to fill this data gap and to determine the degree to which short term velocity variability has changed since observations were last taken at such resolutions.  Our method provides velocity fields at spatial resolutions of 100 m with daily estimates of the three-day average over 11 months.  

\section{A Bayesian method for tracking glacier surface features}
\begin{figure}
\label{fig:flowchart}
\begin{center}
\includegraphics[width=0.65\linewidth]{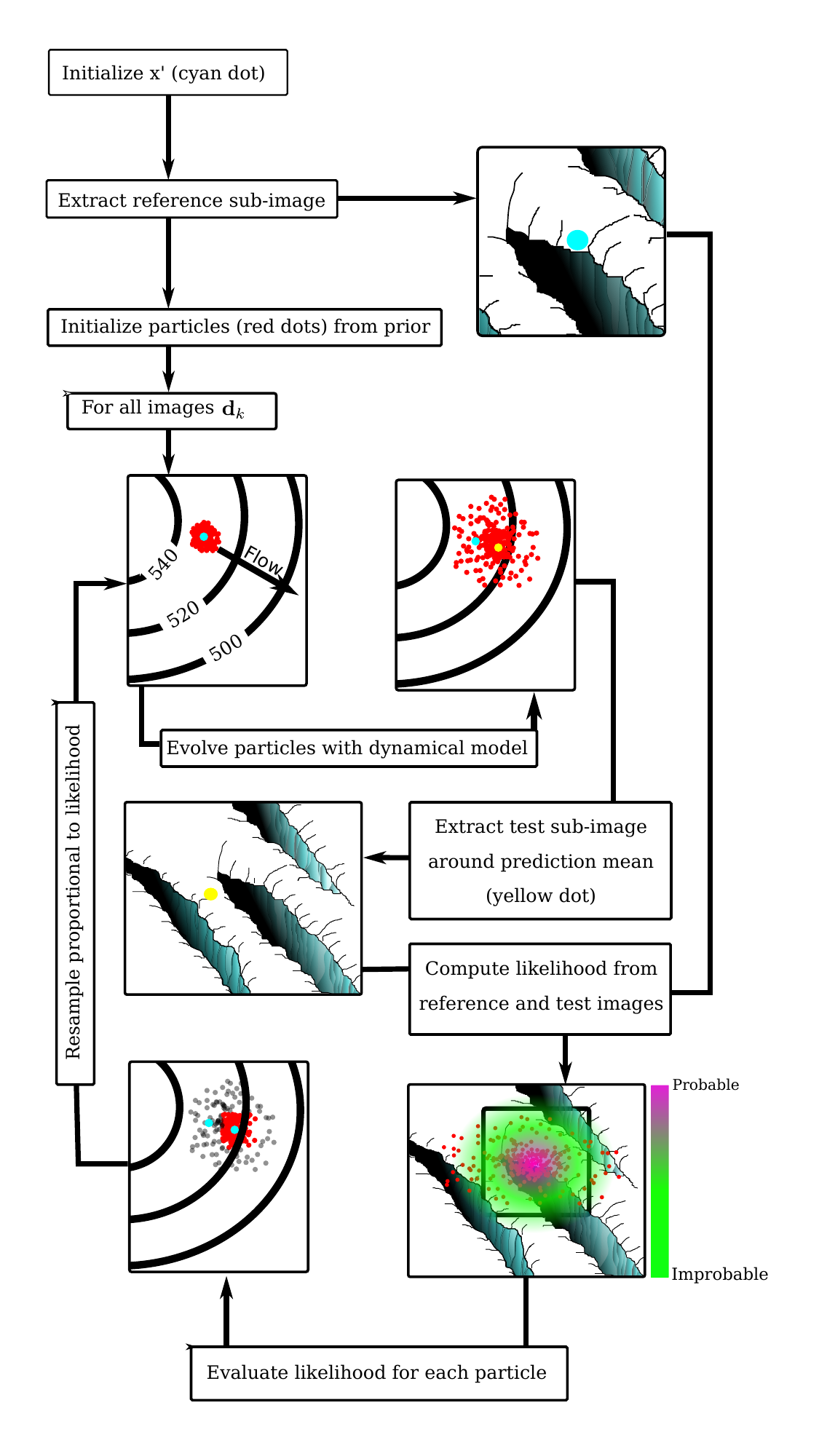}
\end{center}
\caption[Template matching-particle filtering algorithm.]{Graphical depiction of the steps of the hybrid template matching-particle filtering developed in this paper.}
\end{figure}
The problem we seek to solve is as follows: given a set of sequential images of a glacier surface taken from an approximately identical vantage point, find the motion of the glacier and an associated estimate of uncertainty in a spatial (rather than image) frame of reference (See Fig.~\ref{fig:flowchart}).  While methods for tracking features between images are ubiquitous in computer vision, the particular problem of tracking glaciers presents a few notable challenges and requirements.  First, since the images are oblique and we are interested in motion in a spatial frame of reference, the method must be amenable to the use of a projective transformation in some way.  Projection from spatial coordinates to image coordinates is straightforward.  However, the inverse operation of projecting an image onto a landscape that varies in the vertical coordinate is ill-posed because of discontinuities induced by partial occlusion of background elements by foreground elements, particularly as the angle between the camera direction and the surface becomes small.  As such we, wish to avoid this inverse transformation.  Second, glaciers occur outside and in bed weather, so the method must be robust to occasional occlusion and also to changes in lighting.  Third, the method must be able to handle the considerable clutter present on a natural glacier surface.  As a means to address all of these problems, we have developed a method that combines so-called template matching with particle tracking.  

Template matching works by finding the mismatch between a reference sub-image and a larger test sub-image as a function of pixel displacement.  For example, if a test sub-image were  constructed by rigidly translating a reference sub-image by one pixel up and to the right, then the error surface computed by template matching would be minimal for a displacement of one pixel up and to the right.  For natural images, this error surface likely contains multiple minima of various degrees, representing several feasible image offsets of varying certainty.  To improve robustness to changing lighting conditions, we apply local image processing to both reference and test sub-images.  Template matching  utilizes large neighborhoods of data and so is robust to clutter and mismatching compared to similar methods that compare local features \citep[e.g.][]{Lowe:1999}, but it cannot handle large deformation and is somewhat computationally expensive.  Fortunately, neither of these restrictions are too onerous in the context of glaciology, where motion is slow and real-time processing is not required.  

Particle tracking is a method for sequentially updating the probability distribution of the state (primarily position and velocity) of a dynamical model based on a stream of observations (in this case images from a time-lapse camera).  Relevant statistics (such as the mean or maximum likelihood estimator) can then be extracted from the distribution.  The dynamical model that we consider essentially says that a point on the glacier surface moves with nearly constant velocity, subject to small random accelerations typical of glacial velocity variations.  The probability distribution of the state is approximated by a large number of `particles,' each representing a potential state, which are evolved through time based on the stochastic dynamical model.  When observations are available, particles that are unlikely with respect to these observations are culled, while observations that are likely are replicated.  The dynamical model exists entirely in the spatial reference frame.  To determine which particles are likely, they are projected into image space according to a numerical model of the camera, and deemed likely or unlikely  with respect to the error surface computed with template matching.  Note that this allows the algorithm to explore multiple hypotheses, since template matching may produce more than one likely offset.  While many particles are required to fully sample the space of possible velocities, the projection of image offsets back into spatial coordinates is never required.  Additionally, in the absence of observations (or if observations are of low quality due to bad weather), the state evolves according to the dynamical model only. 

We can repeat the above steps for any location desired on the glacier surface.  By applying this method to a grid of points and applying interpolation to the resulting solutions, we thus produce velocity fields.  

\subsection{Glacier motion model}
We use a Lagrangian state-space model (i.e. a set of input variables $\mathbf{m}_k$, a set of output variables $\mathbf{m}_{k+1}$ and a set of first order discrete transfer functions) to represent the motion of trackable features along the surface of a glacier.  The model state variables are the map-plane coordinates $\mathbf{x}$, the map-plane velocities $\mathbf{v}$, the elevation $z$, and a systematic elevation offset from a datum $\delta S$, which form the state vector at the $k$-th time step $\mathbf{m}_k = [\mathbf{x}_k,\mathbf{v}_k,z_k,\delta S_k]$.  We assume that a specified point of interest moves tangent to an imprecisely known glacier surface with its velocity subject to random accelerations, which yields the discrete difference equations
\begin{eqnarray}
\mathbf{x}_{k+1} &=& \mathbf{x}_k + \Delta t \, \mathbf{v}_k + \frac{\Delta t^2}{2}\,\mathbf{a}_k \\
\mathbf{v}_{k+1} &=& \mathbf{v}_k + \Delta t \,\mathbf{a}_k \\
         z_{k+1} &=& S(\mathbf{x}_{k+1}) + \delta S_{k+1} \\
  \delta S_{k+1} &=& \delta S_k + \sigma_z \,\| \mathbf{v}_k \| \,\Delta t.   
\end{eqnarray}
$\mathbf{a}_k$ are normally distributed random accelerations in both horizontal directions
\begin{equation}
\mathbf{a}_k \sim \mathcal{N}(0,\Sigma_{v,k}),
\end{equation}
where $\Sigma_{v,k}$ is a diagonal covariance matrix with entries given by an assumed characteristic variance in glacier velocities.  $S(\mathbf{x}_{k+1})$ is a reference surface elevation field, for example an interpolant to a DEM.  Since errors in this reference surface are likely to be systematic, we assume that uncertainty in the glacier surface evolves as a random walk that depends on how far the particle moves over the DEM, and a characteristic slope of small scale features $\sigma_z$.  We neglect the evolution of the surface elevation due to melting, assuming that this is at least partially accounted for by emergence velocity.  In regions where the ice is flowing quickly the error between the true surface elevation and the reference elevation can evolve quickly, while in regions that are not moving at all, the error should remain approximately constant.  

The initial state $\mathbf{m}_0$ is specified as
\begin{eqnarray}
\mathbf{x}_0 &\sim \mathcal{N}(\mathbf{x}',\Sigma_\mathbf{x}), \\
\mathbf{v}_0 &\sim \mathcal{N}(\mathbf{v}',\Sigma_{\mathbf{v}}), \\
         z_0 &= S(\mathbf{x}_0) + \delta S_{0}, \\
  \delta S_0 &\sim \mathcal{N}(0,\Sigma_S),
\end{eqnarray}
where $\mathbf{x}'$ is the nominal location to be tracked, $\mathbf{v}'$ is an initial guess of the velocity, and the various $\Sigma$ are covariance matrices associated with these initial distributions.  A stochastic state-space model can also be written as a random vector drawn from a distribution conditioned on the previous state, or
\begin{equation}
P(\mathbf{m}_{k}|\mathbf{m}_{k-1}) = \mathcal{N}(\mathcal{F}(\mathbf{m}_{k-1}),\Sigma_{k-1}),
\end{equation}
where $\mathcal{F}$ is the deterministic component of Eqs. 1--4, and $\Sigma_k$ is the covariance matrix associated with the noise.

\subsection{Applying Bayes' Theorem}
The distribution of potential solutions produced by the model described above when used with reasonable estimates of initial distributions and process noise is large.  We wish to determine which of these solutions are likely with respect to a sequence of error-prone observations, in this case the displacement of a characteristic pattern of surface features associated with the the nominal tracked location $\mathbf{x}'$ between oblique images.  Stated more rigorously, we seek the probability distribution of a current state $\mathbf{m}_k$ as constrained by all images up to and including that at the current time $\mathcal{D}_k=\{\mathbf{d}_i:i\in 1,\ldots,k\}$, where $\mathbf{d}_i$ is an image at time $i$.  We restrict our consideration to the case that two assumptions hold, both of which are true in this context.  First, the transition between states must be a Markov process, which is to say that $\mathbf{m}_k$ is independent of all states except the previous one $\mathbf{m}_{k-1}$, or
\begin{equation}
P(\mathbf{m}_k|\mathbf{m}_{k-1},\ldots,\mathbf{m}_0) = P(\mathbf{m}_k|\mathbf{m}_{k-1}).
\end{equation}
This means that the transition between states has no memory.  Second, observations must depend only on the contemporaneous state, and are independent of all other observations and non-contemporaneous states:
\begin{equation}
P(\mathcal{D}|\mathbf{m}_k,\ldots,\mathbf{m}_0) = \prod_{i=1}^n P(\mathbf{d}_i,\mathbf{m}_i).
\end{equation}
Using these assumptions combined with Bayes' Rule \citep{Tarantola:2005} allows us to sequentially update our belief in the state distribution as additional measurements are added:
\begin{eqnarray}
P(\mathbf{m}_k|\mathcal{D}_k) &\propto & P(\mathbf{d}_k|\mathbf{m}_k) \, P(\mathbf{m}_k|\mathcal{D}_{k-1}),
\end{eqnarray}
where $P(\mathbf{d}_k|\mathbf{m}_k)$ is the \emph{likelihood}, which describes how likely it is to observe the current measurement $\mathbf{d}_k$ assuming a state $\mathbf{m}_k$, and $P(\mathbf{m}_k|\mathcal{D}_{k-1})$ is a \emph{prior} distribution that describes how feasible a state is given all previous images but before considering the present image.  Multiplying the likelihood and the prior together yields the \emph{posterior} distribution (to a normalizing constant), which is the probability distribution of states after having considered all available observations.  Note that the posterior distribution (the probability density function of position and velocity of a given point after considering a set of images) can be dominated by either the likelihood or the prior.  In the former case, if an observation is equally likely given any state $\mathbf{m}$ (for example, in the case of complete occlusion of the image by, say, a cloud), then the likelihood is constant and the posterior distribution is only proportional to the prior: the observation has added no new information and reverts to the prior.  Conversely, if observations are very certain and the prior relatively vague, then the posterior distribution will be governed by observations.

The prior distribution is constructed by propagating the posterior distribution at $k-1$ through the state model, which is to say that the best guess for the current state is the fully-constrained previous state updated with the model dynamics.  This turns out to be true, which can be seen by factoring the expression for the prior as
\begin{equation}
\label{eq:kalmogorov}
P(\mathbf{m}_k|\mathcal{D}_{k-1}) = \int P(\mathbf{m}_k|\mathbf{m}_{k-1}) P(\mathbf{m}_{k-1}|\mathcal{D}_{k-1}) \mathrm{d}\mathbf{m}_{k-1},
\end{equation}
which is a form of the Chapman-Kolmogorov forward equation \citep{Doucet:2009}.  In this equation, the first term is the probability distribution of the new state given the old state (or the forward model), and the second term is the posterior distribution from the previous time step.  This equation forms a new distribution by applying the system dynamics to the posterior at the previous time step.  In the case that all distributions are Gaussian and the system dynamics are linear, Eq.~\ref{eq:kalmogorov} can be solved analytically (leading to the well-known Kalman Filter).  However, in this case neither assumption is true due to the non-linear constraint of surface tangent motion, and also due to non-linearity in the process of making observations, described below.  

\subsection{Measurement}
The likelihood can be interpreted as follows: given a known state, what is the probability of a camera recording a given (sub-)image.  The process we adopt here can be summarized as a) project the mean state position into image coordinates, b) extract a sub-image in the neighborhood around the projected point and perform local image processing, and c) compute the sum of squared differences between the sub-image and all other sub-images in the neighborhood, which is interpreted as the scaled logarithm of the likelihood.  

\subsubsection{Specification of a camera model}
Our method relies on possessing an accurate function for projecting coordinates in physical space to coordinates in image space.  We adopt the model of \citet{Claus:2005}, which is specified by camera position, camera orientation, focal length, camera sensor size, and radial and tangential lens distortion.  \emph{A priori}, only the camera location is known, and that with limited precision in this case (namely the precision of a hand-held GPS).  We solve for the remaining parameters by minimizing the misfit between a set of points that are uniquely identifiable in both a digital elevation model and in a reference image (mostly prominent features such as mountain peaks and shoreline outcrops) using Powell's algorithm \citep{Powell:1964}.

\subsubsection{Reference and search sub-images}
We begin by projecting the nominal location of the point we wish to track into image coordinates.  We then find the nearest integer pixel and extract a $m_r \times n_r$ sub-image $T$, which becomes the reference sub-image that we track throughout the period of interest.  As a preprocessing step, we perform a whitened principal components transform to convert the image from RGB to a Z-normalized intensity \citep{Smith:2002}.  We then apply a highpass median filter to highlight edges and partially remove the effect of shadows.

Each time an image becomes available, we compute the weighted mean from $P(\mathbf{m}_k|\mathcal{D}_{k-1})$ (i.e. the prediction step) and project it into image space.  We then extract a test sub-image $I$ with size $m_t \times n_t, m_t>m_r, n_t>n_r$.  The sub-image sizes are found through trial and error, with a tradeoff between feature uniqueness and clutter as sub-image size increases.  We apply the same preprocessing steps with the addition of histogram matching step for each band of the test sub-image, such that the color profile matches that of the reference template.  This helps to ameliorate some of the effects of changing illumination.  

\subsubsection{Computation of the likelihood}
With preprocessed reference and test sub-images in hand, we compute the area-averaged sum of squared differences between the reference template and test template for all possible pixel offsets $u,v$ for which the reference template falls entirely within the test template
\begin{equation}
\ell(u,v) = \frac{1}{m_r n_r} \sum_{u',v'} (T(u',v') - I(u + u',v+v'))^2.
\end{equation}
We then define the likelihood as 
\begin{equation}
\mathcal{L}(u,v) \propto \exp\left(-\frac{\ell(u,v)}{\sigma_\ell^2 + \sigma_m^2}\right),
\end{equation}
where $\sigma_\ell$ is the measurement uncertainty due to illumination changes and deformation, and $\sigma_m$ is the uncertainty due to camera motion in pixels \citep{Nakhmani:2008}.  The deterministic procedure would be to find $\mathrm{argmax}_{u,v} \; \mathcal{L}(u,v)$ (often with a sub-grid parameterization to increase precision) as the (single) measurement.  However, because of the quasi-periodic nature of trackable glacier surface features (namely crevasses) as well as changes in illumination, there are often several peaks in the likelihood, often of comparable magnitude.  Here, we do not try to find this peak, instead retaining the complete likelihood for the update of the posterior distribution.  This allows particles to explore multiple hypotheses for each new image.  Because spurious offsets tend to be inconsistent between images, while the "true" peak is persistent (even if it is not the most probable solution for a given scene), incorrect hypotheses tend to be ephemeral, while good solutions remain probable over multiple images.  

\subsubsection{Camera model correction}
In practice, cameras are not perfectly stable due to changes in temperature, wind, and other unknown factors which conspire to produce small offsets between image pairs \citep[e.g.][]{Harrison:1992}.  To determine this offset, we track a set of points $\{u_g,v_g\}$ that are on land (as opposed to ice or water), and thus assumed to be stationary.  We then compute the maximum likelihood solution with subgrid precision for each control point, and fit a 3 parameter rigid rotation-translation model for each image using the RANSAC algorithm \citep{Fischler:1981} to eliminate outliers induced by occlusions from clouds, errors in choosing the correct motion, etc.  The residual of this fit divided by the number of control points (and penalized for excluding data points) is used as $\sigma_m$, the measurement uncertainty due to camera motion.  When computing the likelihood for non-stationary points, we then add the offset predicted by the rotation-translation model to the projected image coordinates, which minimizes the influence of camera motion.

\subsubsection{Multiple cameras}
\begin{figure}
\begin{centering}
\includegraphics[width=\linewidth]{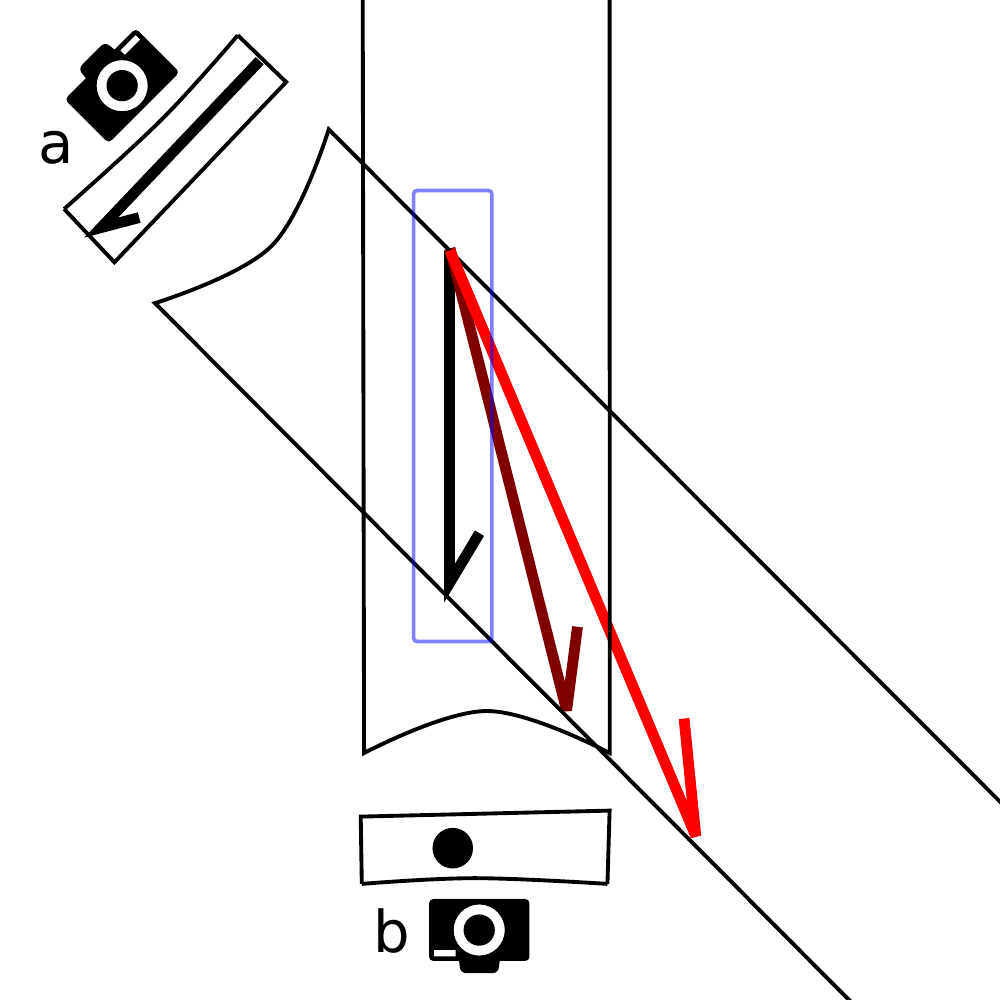}
\end{centering}
\caption[Schematic of two-camera setup]{Schematic illustrating the utility of multiple cameras in oblique flow tracking.  Even if both cameras are insufficient to fully resolve flow direction and magnitude on their own, two cameras operating in tandem typically provide strong constraints on one another.}
\label{fig:camera_geometry}
\end{figure}
Assuming $n_c$ cameras, the above steps can be performed for each, and due to independence, the resulting probabilities can be multiplied.  Because an accounting of observational uncertainty is inherent in the system, the information derived from each camera is properly weighted: cameras far from a given point or with unfavorable geometry are weighted less because likelihood maxima are more diffuse.  

Even cameras that do very little in terms of specifying flow speed on their own can dramatically increase the accuracy of other cameras.  A schematic justification for this is given in Fig.~\ref{fig:camera_geometry}.  In the figure, Camera a successfully captures the component of the flow vector parallel to the image plane.  Unfortunately, there are multiple vectors that can appear identically to the camera.  Though in the real world, this is somewhat offset by the camera being elevated above the surface, this advantage is partially offset by uncertainty in the location of the surface (i.e. DEM errors).  Camera b has a different problem: because the flow vector is normal to the image plane, the camera cannot determine anything about the magnitude of the flow.  However, it does have the advantage of having full knowledge of the flow direction.  Individually, neither of these measurements is very satisfying.  In tandem however, the information from camera b specifies which vector camera a has measured, producing the correct measurement of both magnitude and direction of the offset vector.  As shown in Fig.~\ref{fig:columbia}, the situation at Columbia Glacier is similar to the hypothetical scenario of Fig.~\ref{fig:camera_geometry}.

\subsection{Particle Filtering}
Given that our problem precludes analytical solution, we must instead find an approximate numerical solution.  An effective method for dealing with problems of this type is known variously as sequential Monte Carlo \citep{Doucet:2009}, sequential importance (re-)sampling, bootstrap filtering \citep{Gordon:1993}, particle filtering, or the Condensation Algorithm\citep{Blake:1997}.  We will refer to the method as particle filtering for the remainder of this work.  As the name implies, we rely upon a random sample of feasible states (`particles') that are sequentially updated as new observations become available.  The central assumption is that a probability distribution $P(\mathbf{m})$ can be represented as a weighted set $\{(\mathbf{m}^i,w^i):i\in 1,\ldots,N\}$ of random samples, forming a new probability mass function
\begin{equation} 
\label{eq:discrete_distribution}
P(\mathbf{m}) \approx \sum_{i=1}^N w^i \delta(\mathbf{m} - \mathbf{m}^i),
\end{equation}    
where $\delta(\cdot)$ is the Dirac delta function, and $N$ is the number of random samples.  As $N$ increases, the quality of the approximation increases.  The random samples are drawn from a proposal density $q(\mathbf{m}_k^i|\mathbf{m}_{k-1}^i,\mathbf{d}_k)$, and the weights are computed as
\begin{equation}
w_k^i \propto w_{k-1}^i\frac{P(\mathbf{d}_k|\mathbf{m}_k^i) P(\mathbf{m}_k^i|\mathbf{m}_{k-1}^i)}{q(\mathbf{m}_k^i|\mathbf{m}_{k-1}^i,\mathbf{d_k})},
\end{equation}
which are subsequently normalized.  The proposal distribution is arbitrary, but some choices are better than others with respect to capturing the posterior distribution with a minimum number of samples.  The practical and intuitive choice is that the proposal distribution should be the prior distribution at time $k$
\begin{equation}
q(\mathbf{m}_k^i|\mathbf{m}_{k-1}^i,\mathbf{z}_k) = P(\mathbf{m}_k^i|\mathbf{m}_{k-1}).
\end{equation}

While this method ostensibly captures the posterior distribution given enough particles, the diffusive nature of the state equations means that eventually very few particles will be probable with respect to observations.  We overcome this by resampling at each time step from the samples $\mathbf{m}^i$ with probability given by $w^i$ using a systematic resampling method \citep{Carpenter:1999}.  Resampling produces the same distribution with particles now distributed proportional to the weights, which are then reset to $1/N$.  Weights are now simply proportional to the likelihood
\begin{equation}
w_k^i \propto P(\mathbf{d}_k|\mathbf{m}_k^i).
\end{equation}
The resulting distribution converges to the true posterior probability 
\begin{equation}
P(\mathbf{m}_k|\mathcal{D}_k) = \lim_{N\to \infty} \sum_{i=1}^N w^i_k \delta(\mathbf{m}_k - \mathbf{m}_k^i),
\end{equation}
proof of which can be found in \citet{Blake:1997}.

\section{Application to Columbia Glacier}
We apply the above algorithm to two cameras, dubbed AK01 and AK10 (Fig.~\ref{fig:columbia}) from 2013.06.10 through 2013.09.25 and from 2013.11.06 through 2014.4.30, during which both cameras operated continuously at 20 minute intervals.  AK01, a Nikon D200, recorded 5.8 megapixel images in JPEG format at quality level 99.  AK10, a newer D200, recorded 10 megapixel images at quality level 92.  The break between camera epochs naturally leads to a `summer' and `winter' record, and we divide our analysis along those lines.  

To specify the surface elevation $S(\mathbf{x},t)$, we linearly interpolate (in time) between the nearest two members of a set of 10m-resolution digital elevation models derived from the TanDEM-X satellite \citep{Vijay:2017}, to which we fit a 3rd order spline for sub-pixel interpolation.  DEMs at this level of precision capture transient crevasse features.  To account for this, we sequentially apply a maximum filter followed by a Gaussian smoother over the glacierized area, each with a 30m kernel, which has the practical effect of `filling' crevasses.  More complex approaches to surface processing are possible \citep{Messerli:2014}, but experimentation has shown the results to be relatively insensitive to the smoothing method.

We specify initial locations of points to track as the vertices of a grid with 100m spacing in both map-plane coordinates, so long as those points are within the field of view of both cameras, and the elevation is more than 20 m above sea level (Vectors in Fig.~\ref{fig:tsx_comparison} correspond to these points).  We start the algorithm for each day in the record period at noon local time (20:00 UTC), and track for 3 days.  The algorithm works equally well running backwards in time, so we also track the point backwards starting at the final image of the forward run.  We take the mean of the two resulting velocities weighted by the inverse of the Frobenius norm of the sample covariance (i.e. cases in which the last or first image has bad weather has high covariance so contributes little to the mean).  

We must make a few choices regarding process noise, and these are mostly informed by heuristics.  We assume that random accelerations have a standard deviation of approximately 2 m d$^{-2}$ in both directions.  This is based on the characteristic velocity variations observed from high resolution surveys \citep{Meier:1993,Krimmel:1986}, and also so that the model has the potential to capture the abrupt slowdowns sometimes observed at Columbia Glacier in the fall (O'Neel, unpublished data).  We specify the standard deviation of local slope as $\sigma_z=0.1$. 

We use a square reference image with size $m_r=n_r=15$ pixels, and a test image size of $m_t=n_t=25$ pixels, which assumes a maximum search distance of 10 pixels.  Note that this is a form of prior information that we are explicitly introducing into the results: we assume that the probability that a particle moves more than 10 pixels in image space is zero.  In practice, this turns out to be a reasonable assumption, and serves to limit spurious correlations and relieve computational effort.  We assume $\sigma_t=0.25$,  which implies that a correlation peak is localized to within a 1 pixel range with $2\sigma$ credibility.  Values of $\sigma_m$ vary between images, but when matching two images in good weather under similar illumination, the value is close to zero, while unfavorable conditions where many points are occluded will produce uncertainties of greater than $\sigma_m=5$ pixels, which effectively means that the likelihood is uniform across the test image.  

We approximate the probability distribution at each point we wish to track with $N=3000$ samples.  This is probably overkill; however, since our application does not need to be run in real time, the improved convergence associated with using many particles is worth the increased computational costs.  Since each sample point is independent, the code is naively parallelizable.  The lion's share of memory is taken up by storing many high-resolution images.  We utilize a shared read-only memory structure, so that computation for each point in the grid can draw upon the same location in memory.  Otherwise each grid point is independent.  On a laptop with 8 cores, processing each scene requires around 150 s, depending on the time of year (with maximum and minimum computational times falling on the solstice due to the abundance or dearth of usable images).  The processing time would decrease linearly with more cores.   

The result is a three-day running average for each initial point at daily resolution, encoded in the form of a collection of sample from which we can draw statistics.  While the converged velocity distributions are not normal, they are sufficiently close that they can be described by a mean and covariance matrix.  To eliminate outliers and to smooth the resulting fields, we replace each velocity and covariance component at each point with the medians of neighboring points within 150m.  Henceforth, when we refer to velocities (and uncertainties), unless otherwise specified, we refer to the distribution mean (and covariance) smoothed in this way. 

\section{Results}
\subsection{Pointwise velocity evolution}
\begin{figure*}
\begin{center}
\includegraphics[width=0.8\linewidth]{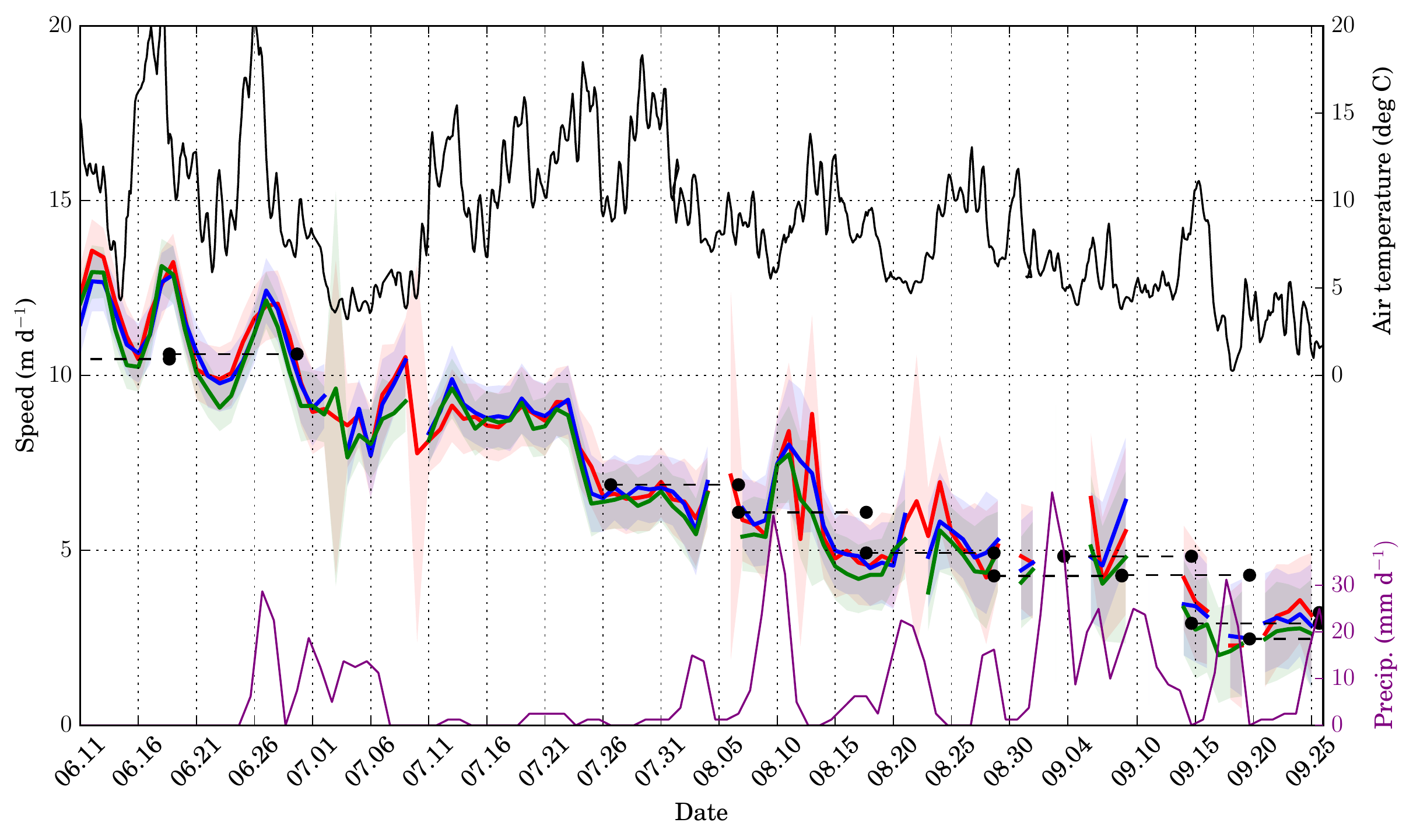}
\includegraphics[width=0.8\linewidth]{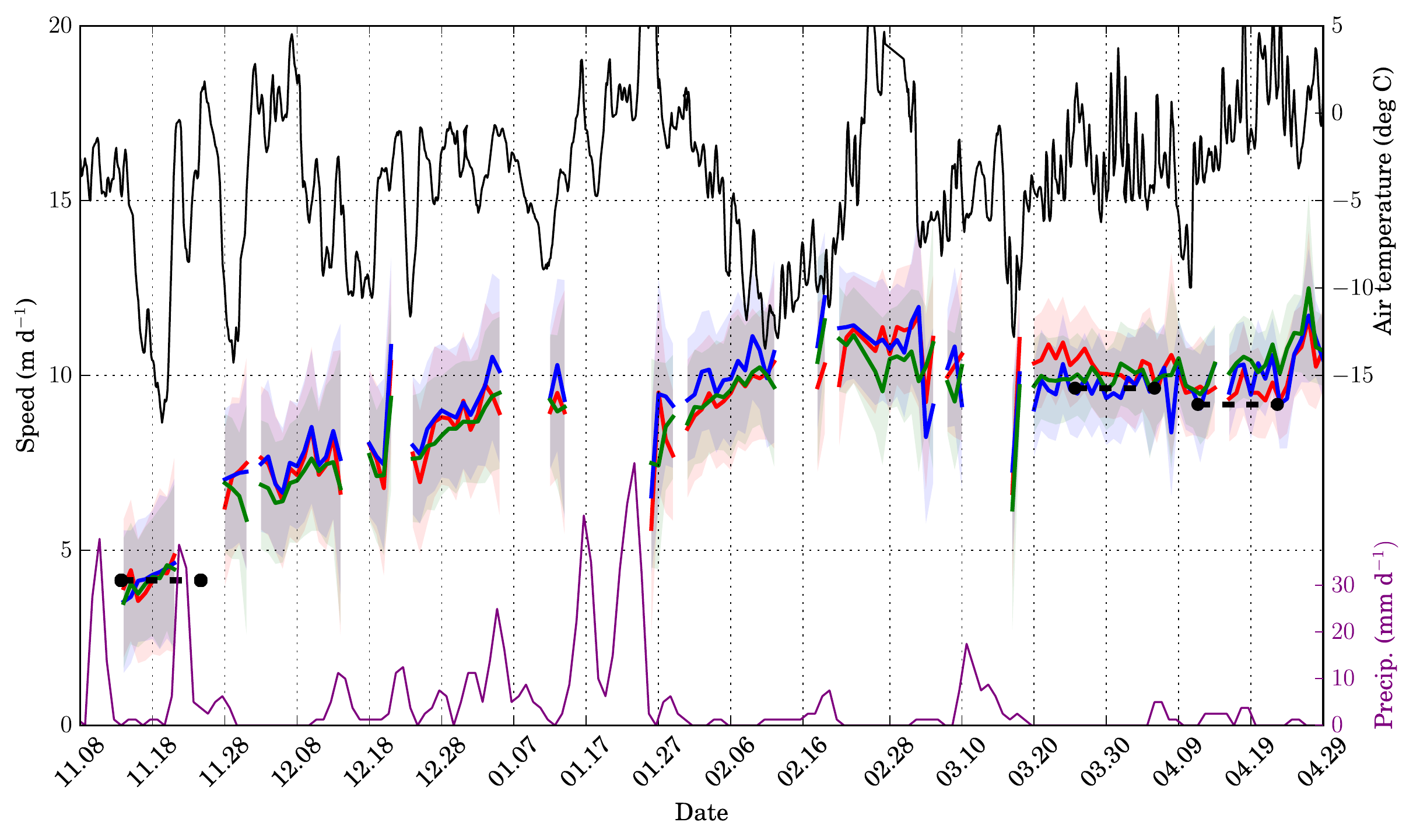}
\end{center}
\caption[Time series of pointwise ice speed.]{Time series of ice speed at points corresponding to line color in Fig.~\ref{fig:tsx_comparison} in summer 2013 (top) and winter 2013-2014 (bottom).  Black barbells are TSX speeds over the time periods indicated by the line endpoints.  The black line in the upper portion of the plot is air temperature at a weather station approximately 5 km upstream from the Columbia Glacier terminus and at an elevation of \~1000 m, while the purple line is precipitation rate recorded in the village of Tatitlek, near the mouth of Columbia Bay.}
\label{fig:time_series}
\end{figure*}
The velocity of the near-terminus shows strong temporal variability over seasonal and sub-seasonal time scales (Fig.~\ref{fig:time_series}).  A point near the terminus attains a velocity maximum of nearly 15 m d$^{-1}$ on 2013.06.12, near the start of the record, before decreasing to a minimum of 3 m d$^{-1}$ in late September.   During the summer, the glacier also shows a variety of distinct speed-ups.  These speed-ups are correlated with warm periods during June, but this correlation is only weakly evident by later in the season.  Large rainstorms in August and September produce speedups as well, each lasting around three days.  Due to the tendency for the glacier to be occluded during these weather events, they are also the most uncertain.  

Following the fall velocity minimum, Columbia Glacier begins to accelerate in mid-November.  The speed then remains relatively consistent throughout the winter, with a maximum velocity of around 12 m d$^{-1}$ in February before falling to 10 m d$^{-1}$ throughout March and April.  During the winter, the glacier is non-responsive to variations in precipitation in temperature, presumably because temperatures are too cold for liquid water to be present at the glacier surface in large volumes (though plotted temperatures are recorded at 1000 m and are usually several degrees lower than those at the elevation of the glacier terminus).  Note that during the winter months, the general level of uncertainty in speed is higher.  This is because fewer usable images exist at this time due to less daylight: less data means the flow is less precisely resolved.  

\subsection{Spatial patterns of velocity change}
\begin{figure*}
\includegraphics[width=\linewidth]{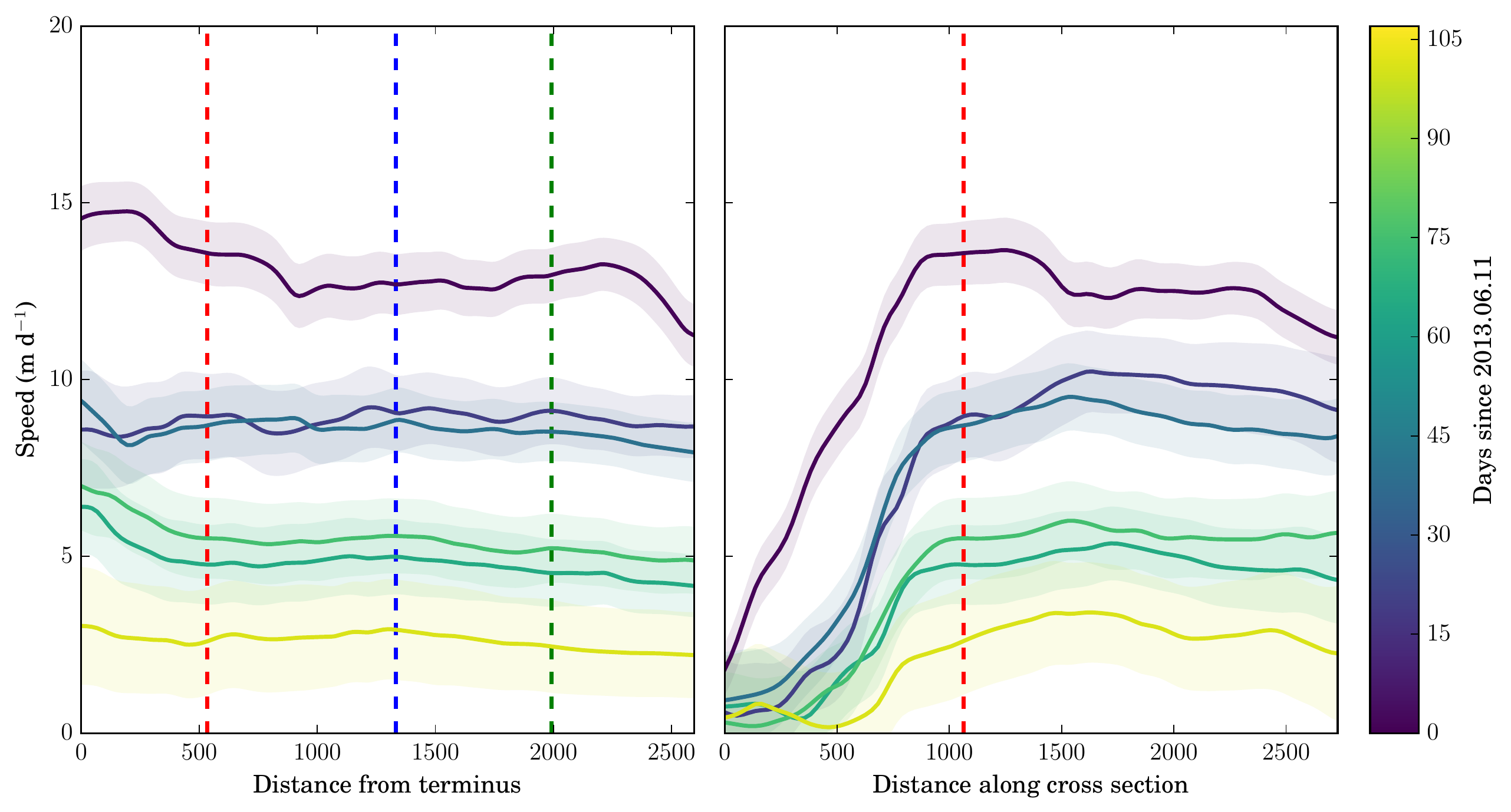}
\caption[Along- and across-flow speed profiles.]{Along-flow (left) and across-flow (right) speed profiles during summer 2013, with colored lines representing the location at which the time-series in Fig.~\ref{fig:time_series} were extracted.  Note the relative homogeneity of speeds in the along-flow direction, which are mostly constant except for a ephemeral acceleration within 500 m of the terminus.  In the cross-sections, note the activation and deactivation of marginal ice between 0 and 500 m.}
\label{fig:profiles}
\end{figure*}
In the nearest 2.5 km from the terminus along the centerline, Columbia Glacier shows relatively small spatial variability in speed (Fig.~\ref{fig:profiles}a).  Nonetheless, because the baseline speed at Columbia Glacier is so fast, these small variations are still sufficient to produce large strains and the evident extensive crevassing.  A slight acceleration in the lowest 500 m seems to be transient, and we could detect no specific reason for why this occurs.  Changes in the stress regime due to individual calving events could be a factor, but previous work indicating velocity changes over this scale have been for a floating terminus \citep{Murray:2015,Ahn:2010}, whereas Columbia Glacier's terminus is grounded over the period in question.

Looking at the glacier in cross-section, a migrating shear margin is evident.  Ice near the edge of the glacier is activated or deactivated based on fast flow in the center.  Extrapolating from the edge of the data, at its fastest rate in mid-June, the entire glacier width is in motion.  Later in the year, when centerline speeds have dropped to 20\% of maximum, a nearly 1000 m wide strip of marginal ice become stagnant.   

\subsection{Validation}
\begin{figure*}
\includegraphics[width=\linewidth]{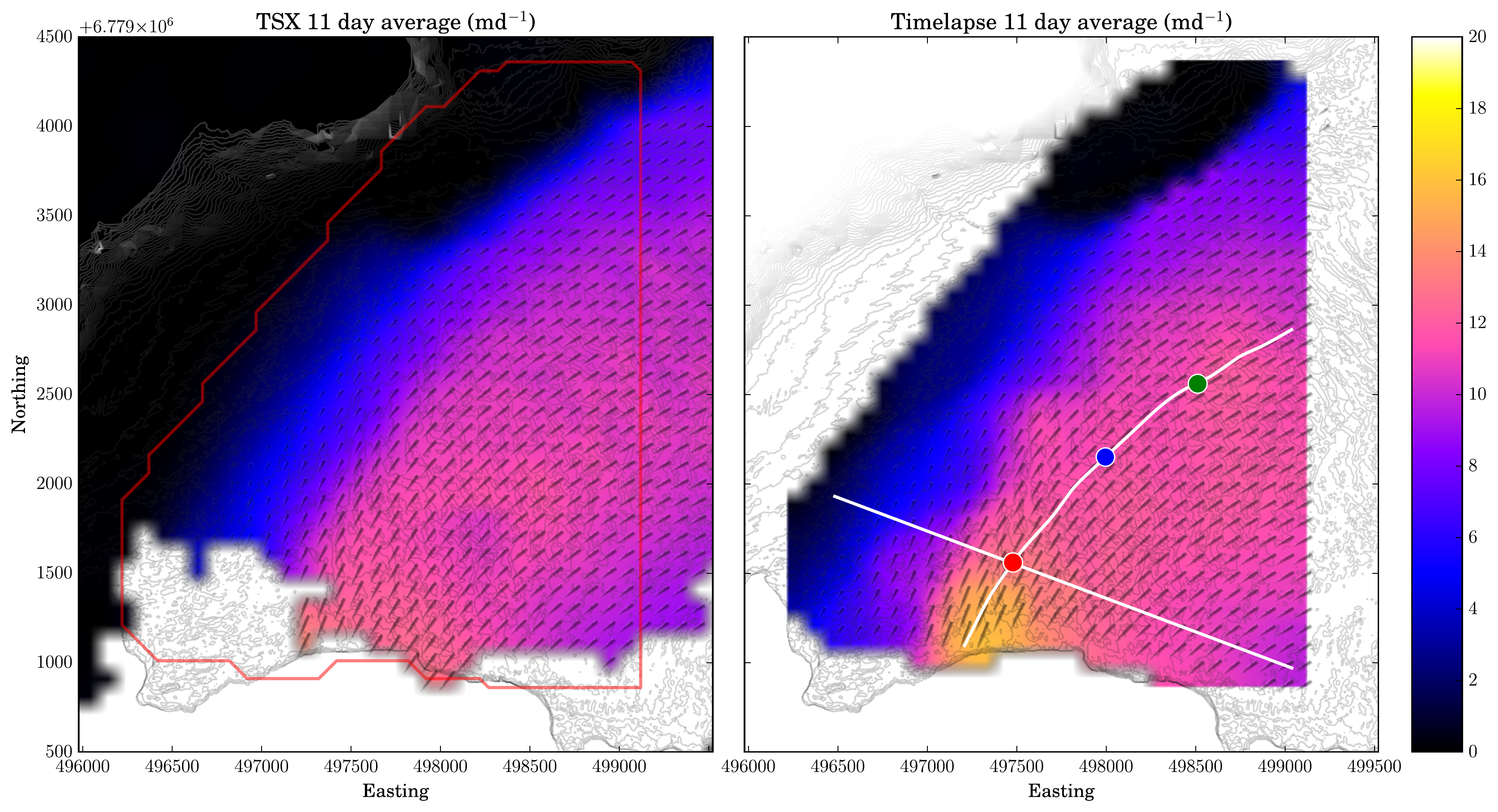}
\caption[Comparison of velocity fields computed using SAR and time-lapse methods.]{Comparison of a TSX-derived velocity field covering 2013.06.10 through 2013.06.21 (left) to ones produced using this method averaged over the same period (right).  The colored dots are the locations at which the lines in Fig.~\ref{fig:time_series} are extracted, and the white lines correspond to the long profile and cross-section of Fig.~\ref{fig:profiles}.  The distinct shear margin in the northern part of the field corresponds to the one visible in Fig.~\ref{fig:images}.}
\label{fig:tsx_comparison}
\end{figure*}
\begin{figure}
\includegraphics[width=\linewidth]{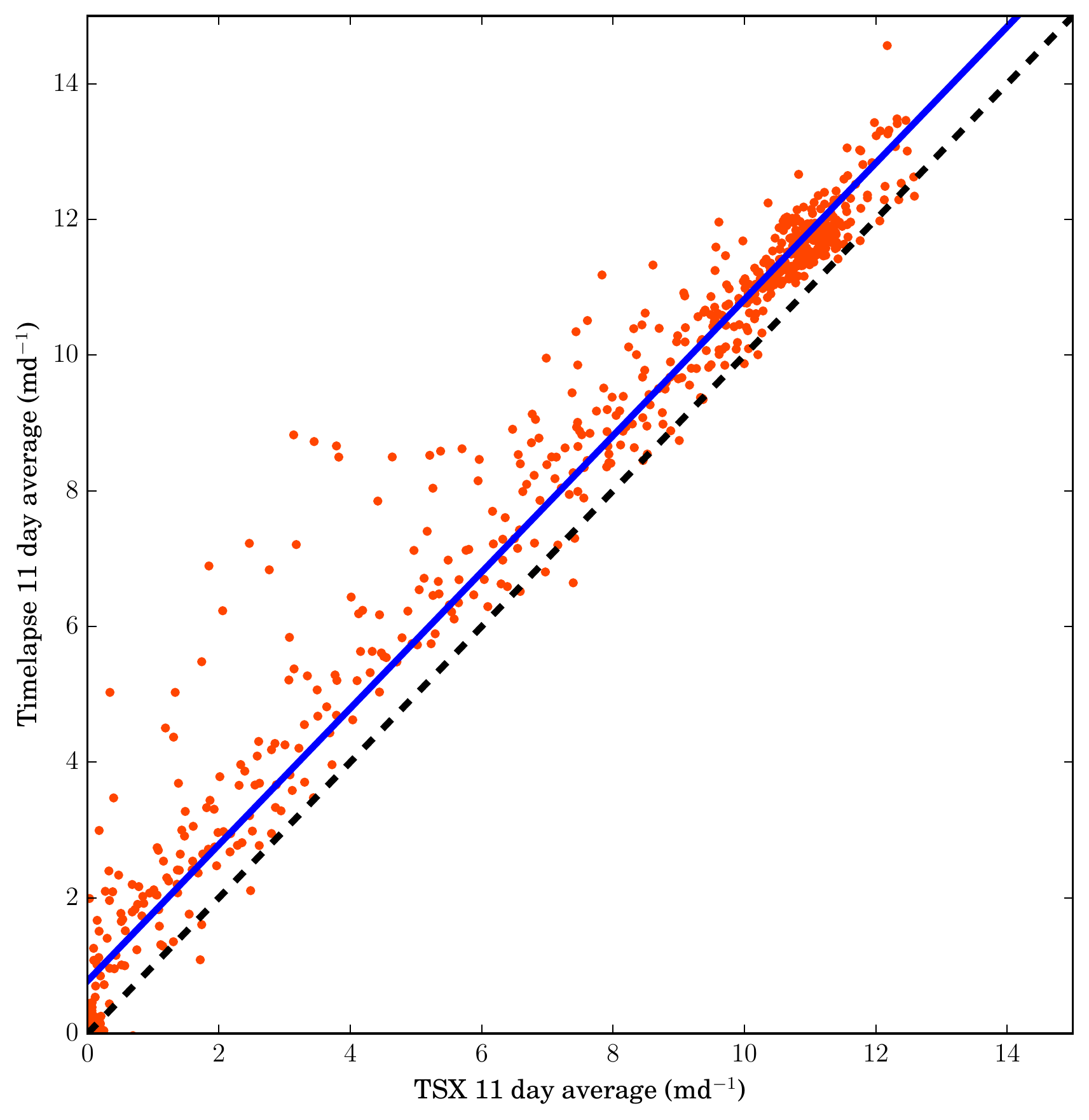}
\caption[Pointwise correspondence between TerraSAR-X and particle filtered velocities.]{The correspondence between the speeds of Fig.~\ref{fig:tsx_comparison}.  The black dashed line is 1:1.  The blue line is the best fit line from a robust regression, with slope close to 1 and a bias of 0.7 m d$^{-1}$.  Other epochs show a similar level of agreement.}
\label{fig:tsx_line}
\end{figure}
It is necessary to confirm that the method is producing results at least consistent with other contemporaneous observations, if not with reality.  To this end, we compare the time average of the daily velocity computed with the particle filtering method over the same epoch as a velocity field derived from the radar interferometer TerraSAR-X (TSX) \citep{Joughin:2010}.  Fig.~\ref{fig:tsx_comparison} shows a TSX velocity field taken for 2013.06.10 through 2013.06.21 and the stacked velocity field derived for time-lapse for the same period.  The qualitative agreement between the two fields is excellent, with the shear margin in the northeast corner captured in both, as well as the locations of fast flow.

The correspondence between speeds produced by both methods is shown in Fig.~\ref{fig:tsx_line}.  A robust linear regression gives a slope of nearly unity, $r^2=0.97$, and a bias of 0.7 m d$^{-1}$, with our particle filtering method producing the faster speeds.  The average standard deviation in the time-lapse speed over the averaging period is $\sigma_V=1.7$m d$^{-1}$, and computed standard deviations encompass the TSX velocities in nearly all cases.  Reported uncertainties for TSX range between 1--10m d$^{-1}$, and similarly encompass the time-lapse speeds.  Results are similar for all epochs for which optical conditions allow for good velocity solutions.  The good fidelity between datasets lends confidence to the method for subsequent interpretation.

It is also interesting to examine the differences between the two velocity fields.  The particle filtering method shows a patch of fast ice in the $\sim$200 m upstream from the terminus.  Given the high velocities in this region, it would make sense that this speedup is not captured by SAR: trackable surface features would be calved off in the eleven-day interval between satellite images, while the higher temporal resolution offered by time-lapse can track features much nearer the calving front. 

As an additional validation measure, TSX velocities, when they exist, are overlain in Fig.~\ref{fig:time_series}.  In general, there is excellent agreement, but with our method showing considerably more variance than can be captured in the 11-day averages of TSX.  This variance is due to the additional resolution granted by analyzing time-lapse imagery, which allows us to capture speed changes associated with short term changes in water input or perhaps individual calving events \citep{Ahn:2010}. 

\section{Discussion}
\subsection{Implications for the subglacial hydrologic system}
Our results strongly support the conclusions of previous studies that suggest that annual and sub-annual velocity variations are driven by the evolution of the subglacial hydrologic system in response to changes in external input in the form of either melt or rain \citep{Kamb:1993,Fahnestock:1991,Vijay:2017}.  The seasonality evident in our results is consistent with observations throughout Columbia's retreat history, and we support the following explanation for it.

During mid-winter, external water input to the glacier base is likely small, particularly given the availability of copious surface snow to absorb moisture, even in the few cases where winter temperatures climb above freezing.  The hydrologic system at this time is in a dormant state where basal water pressures are governed only by basal melt, and the ocean, although remnant water from the ongoing drainage of firn may also contribute.  The tidewater influence is important: baseline hydraulic head is still a significant fraction of flotation, and this allows the glacier to maintain a minimum flow speed an order of magnitude greater than the maximum speeds of most land-terminating glaciers.  Nonetheless, because there is little water flux, efficient drainage elements such as Rothlisberger channels \citep{Rothlisberger:1972} likely cannot persist because there is insufficient turbulent heat dissipation to maintain them.  Thus we echo \citet{Kamb:1993} and hypothesize that the winter state is characterized by an inefficient drainage system that lead to moderately high water pressure and flow speed.  We would anticipate that velocities during winter would be sensitive to the odd winter water pulse (e.g. a so-called pineapple express where sub-tropical air and moisture are transported to northerly latitudes).  However, such events in 2013--2014 were invariably accompanied by severe camera occlusion, and we could not discern velocities during this time period with any certainty.  Nonetheless, there is some evidence of short lived slow-downs following large precipitation events in January and March.  

At the onset of melting in the spring, the subglacial drainage system is overwhelmed by the availability of water.  Since the drainage system is still inefficient, additional input must be accommodated by an increase in water pressure, which leads to fast flow.  This sensitivity is shown clearly during June of 2013 (See Fig.~\ref{fig:time_series}a), during which the velocity and air temperature (a proxy for melt) are highly correlated.  This strong correlation is short-lived, however.  We propose that as the subglacial drainage system develops, more water is required to maintain high velocities.  The excess water from subsequent warm spells loses its impact come July, with speeds decreasing throughout the summer.  Interestingly, this slowdown does not appear to occur gradually; rather it decreases in discrete steps, the most distinct of which is seen around 2013.07.24.  This event is particularly interesting because there seems to be no discernible external forcing, although it is worth noting that changes in melt rates may be independent of temperature \citep{Sicart:2008}.  Temperatures had been high but not anomalous, and precipitation was minimal.  It seems possible that this could be triggered by a sudden change in the transmissivity of the subglacial hydrologic system, such as the confluence of two large Rothlisberger channels, but this is purely speculative.  

Throughout late summer, major rainfall events drive short-lived ($\sim$ 3 day) speed peaks, though it is difficult to be sure of the magnitude of these peaks due to occlusion of the camera by the same precipitation that produced the speed-up.  However, two events in August were observed with reasonable precision.  A simple calculation of melt using a temperature index model \citep{Hock:2003} implies that meltwater and rainwater flux during these large rain events were both around 40 mm of water equivalent per day, though this likely underestimates melt due to latent heat transfer from precipitation.  This combined influx produced speed-ups that were slightly lower in magnitude to the purely melt-induced speed-ups of the early season.  The modest speedups relative to enhanced influx indicates that the subglacial drainage system has evolved to efficiently transmit the extra water.  In the long term, rain events may enhance the process of seasonal slowdown.  After the rain event of 2013.08.08-2013.08.12, velocities dropped by 30\% compared to the pre-storm average, though as discussed above, such a drop is not necessarily associated with external forcing.  In any case, these slowdowns are consistent between tidewater glacier systems, but not typically evident in terrestrially terminating glaciers.  This may be caused by the lower limit on water pressure imposed by the marine margin, which keeps conduit pressure high enough to resist fast closure \citep{Podrasky:2012}.   

In early fall, the situation is opposite to that of mid-winter.  The drainage system is well-developed, but as temperatures drop and rain transitions to snow, influx becomes small.  What little water is available is drained through an efficient hydrologic system kept open by the background water pressure imposed by sea level.  Eventually, as creep closure destroys this efficient drainage system and slow flow keeps subglacial storage capacity to a minimum \citep{Bartholomaus:2011}, water pressure and speeds recovers to winter levels, which seems to be complete by approximately January.  

\subsection{Improvements and extensions}
As with all numerical methods, there are a number of ways in which the algorithm's performance can be improved.  First, the problem of image registration and co-registration yields a source of systematic uncertainty that is difficult to quantify.  Cameras tend to be placed such that glaciers take up most of the field of view.  Unfortunately, glaciers are useless for the purpose of establishing ground control for the calibration of camera models.  Instead, ground control is established by attempting to match features in error-prone DEMs to recognizable features in images that are typically very far (e.g. mountain peaks) or very near.  Variable atmospheric distortion compounds these errors.  Additionally, manual digitization of ground control points is labor-intensive and subject to picking errors.

A potential method for circumventing these difficulties is to utilize automated detection of horizons and other strong image edges, and minimize the difference between these and horizons computed from a digital elevation model.  Such a method could be used simultaneously to find static camera parameters such as focal length and lens distortion and also time-varying camera orientation \citep{Baboud:2011}, without any human intervention.  

Second, an adaptive procedure for selecting reference images could be included.  As it stands, we track a sub-image extracted from the first image in a sequence, irrespective of the fact that it could be occluded, which obviously leads to failure (fortunately this failure is reflected in uncertainty estimates).  Instead, we could compare the statistical properties of each image in a sequence in order to find one that is of good quality for tracking.  The method would still fail if too many images in a sequence were occluded, but it may help fill some of the gaps evident in Fig.~\ref{fig:time_series}.  However, we would also not know the location of the selected sub-image at the beginning of the sequence, and a more carefully selected prior would be needed to ensure that the image correlation procedure does not become lost.  One potential choice is to use the posterior from the last epoch in which the velocity fell below a certain error threshold, or to initialize the algorithm with SAR-derived velocities.  

Finally, while the results presented here rely on the use of two cameras in order to properly specify flow directions, those directions deviated little throughout the observation period despite large changes in speed.  This suggests that we could extend the observational record to include epochs in which only one camera was active by specifying mean flow direction as a strong prior on the computed velocities.  While flow directions are likely to change substantially in the long term, this could be extremely useful for filling gaps, such as the one between 2013.09.25 and 2013.11.06 that exists in the data presented here, wherein AK10 failed while the other camera continued to take images uninterrupted.  


\subsection{Conclusions}
We developed a probabilistic method for tracking glacier surface motion based on time-lapse imagery.  The method operates by evolving a set of particles according to a stochastic dynamical model, while culling particles that are improbable and reproducing probable ones, with the likelihood determined by computing the sum of squared differences between a reference image and test image.  The resulting set of solutions converges to the true posterior distribution of glacier velocity at a given point.  The method is robust to occlusion and false matching, provides rigorous uncertainty, and easily accommodates the refinement of velocity measurements with the use of multiple cameras.  

We apply the developed method to just under a year's worth of images collected by two cameras near the terminus of Columbia Glacier between 2013.06 and 2014.05.  Based on image geometry and distances, the method was able to extract three day running-average velocities over all time periods during which the glacier surface was visible.  To ensure that the resulting velocity fields were valid, we compared TerraSar-X derived velocities to quantities computed with this method and averaged over the same temporal footprint, finding excellent agreement over the entire record.  

At seasonal time scales, our findings mirror those of previous workers in showing that Columbia Glacier transitions between a winter state characterized by moderate velocities, to an early summer speed-up, to a fall slowdown, in which velocities drop to well below their winter state before eventually recovering in early winter.  Our method resolves velocity correlations with melt and rainfall events, though the glacier's sensitivity to these events appears seasonal: during the spring, before an efficient drainage network has developed the glacier sees strong melt- and rainwater induced variability, while in the fall the system responds very little to these forcings.  The velocity fields produced here may help to constrain future simulations of tidewater glacier hydrology and the resulting changes in ice dynamics. 

\section*{Acknowledgements}
Brinkerhoff was supported by a fellowship from the National Science Foundation (NSF-DGE1242789) and the National Science Foundation Graduate Research Internship Program.  We thank Ethan Welty for insightful discussion.  We thank Martin Truffer and Andy Aschwanden for comments which improved the manuscript.

\bibliography{tracker_bib}   
\bibliographystyle{agufull08}

\end{document}